\documentclass{article}

\usepackage{fullpage}

\usepackage{amsmath,amsfonts,amsthm,amssymb}
\usepackage{csquotes}

\usepackage{stmaryrd}
\usepackage{wrapfig}
\usepackage{multirow}
\usepackage[mathscr]{euscript}
\usepackage[shortlabels]{enumitem}
\setlist{nolistsep}
\usepackage{array}
\usepackage{rotating}
\usepackage{diagbox}
\usepackage{float}

\usepackage[utf8]{inputenc} 
\usepackage[T1]{fontenc}    
\usepackage{hyperref}       
\usepackage{url}            
\usepackage{booktabs}       
\usepackage{amsfonts}       
\usepackage{nicefrac}       
\usepackage{microtype}      

\usepackage{graphicx, xcolor} 
\usepackage{subfigure} 

\usepackage{caption}

\usepackage{verbatim}

\usepackage{algorithm}
\usepackage{algorithmic}

\newtheorem{theorem}{Theorem}

\newcommand{\vF}{\mathbf{F}} 
  
\newcommand{\vh}{\mathbf{h}}
\newcommand{\vx}{\mathbf{x}}
\newcommand{\vb}{\mathbf{b}}

\newcommand{\vg}{\mathbf{g}}

\newcommand{\vv}{\mathbf{v}}

\newcommand{\vz}{\mathbf{z}}

\newcommand{\vzero}{\mathbf{0}}

\newcommand{\clip}{\mbox{clip}}

\newcommand{\hy}{\tilde{y}}

\DeclareMathOperator*{\argmin}{arg\,min}
\DeclareMathOperator*{\argmax}{arg\,max}
\DeclareMathOperator{\sgn}{sgn}

\newcommand{\RR}{\mathbb{R}}      

\newcommand{\vnorm}[1]{\left\lVert#1\right\rVert} 
\newcommand{\ifn}[1]{\mathbf{1}\left(#1\right)} 
\newcommand{\abs}[1]{\left| #1 \right|}

\newcommand{\ip}[2]{\left\langle #1, #2 \right\rangle}

\newcommand{\algname}{\textsc{Marvin}} %

\newcommand{\hcname}{\textsc{HedgeMower}}

\newcommand{\cH}{\mathcal{H}}

\newcommand{\lrp}[1]{\left(#1\right)}
\newcommand{\lrb}[1]{\left[#1\right]}
\newcommand{\lrsetb}[1]{\left\{#1\right\}}

\newcommand{\authcmt}[2]{\textcolor{#1}{#2}}

\newcommand{\toinsert}[1]{\authcmt{cyan}{[#1]}}

\newcolumntype{M}[1]{>{\centering\arraybackslash}m{#1}}
\makeatletter
\newcommand{\thickhline}{%
    \noalign {\ifnum 0=`}\fi \hrule height 1pt
    \futurelet \reserved@a \@xhline
}
\makeatother

\makeatletter
\def\blfootnote{\xdef\@thefnmark{}\@footnotetext}
\makeatother

\title{Muffled Semi-Supervised Learning}

\author{
  Akshay Balsubramani \\
  University of California, San Diego\\
  \url{abalsubr@ucsd.edu}
  \and
  Yoav Freund \\
  University of California, San Diego\\
  \url{yfreund@ucsd.edu}
}

\begin{document}

\maketitle

\begin{abstract} 
We explore a novel approach to semi-supervised learning. This approach is contrary to the common approach in that the unlabeled examples serve to "muffle," rather than enhance, the guidance provided by the labeled examples. We provide several variants of the basic algorithm and show experimentally that they can achieve significantly higher AUC than boosted trees, random forests and logistic regression when unlabeled examples are available. 
\end{abstract} 

\section{Introduction}

The \emph{boosting} approach to learning binary classifiers is to construct a
weighted-majority ensemble of them by incrementally adding base classifiers(\cite{SF12}). 
This process is guided by a potential function that is an upper bound on the training error. 
The weights assigned to the training examples communicate the gradient of the potential function to the base learner.

Like other supervised learning algorithms, boosting algorithms require a sufficiently large labeled training set in order to produce an accurate classifier. 
Such algorithms do not make use of \emph{unlabeled} data, which is typically much more abundant.

On the other hand, semi-supervised learning approaches~(\cite{CSZ06})
attempt to use both labeled and unlabeled training examples. The basic
idea in many approaches is to augment the labeled set by inferring the
label of unlabeled examples from their labeled neighbors in some way. 
Such inference uses the "hallucinatory labels" on the unlabeled data that tend to agree with the labeled ones (e.g. \cite{BNS06}). 
In this paper, we present a semi-supervised learning approach that uses the opposite strategy. 
Instead of using the unlabeled examples to \emph{enhance} the labeled examples, we use the unlabeled examples to \emph{muffle}
the effect of the labeled examples, and hallucinate labels which tend to oppose the labeled ones. 

This strategy arises from a transductive inference approach, 
assuming that the labeled and unlabeled examples are drawn from the same distribution. 
To create labeled and unlabeled training sets, 
the label of each example is either exposed (labeled example) or left hidden (unlabeled example) independently at random. 
The task of learning in this scenario is to accurately predict the label of the unlabeled examples. 
This task is significantly easier (\cite{V82}) than the task of standard (inductive) learning, which is to generate a rule 
that will accurately predict on {\em any} as-yet-unseen examples drawn from the same distribution as the training set.

\begin{wrapfigure}{r}{0.5\textwidth}
  \begin{center}
    \includegraphics[width=0.4\textwidth]{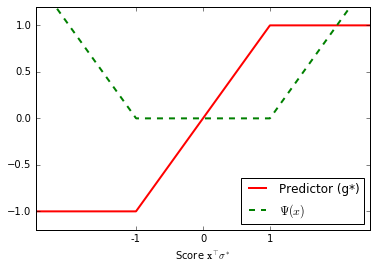}
  \end{center}
  \caption{Potential well and prediction on an unlabeled example, as a function of its score.}
  \label{fig:predscorefunc}
\end{wrapfigure}

In this paper, we devise algorithms which build empirically on recent work of
\cite{BF15}.  That paper directly considers the \emph{test} error
based on the error rates of the ensemble classifiers (described in
Section \ref{sec:boundtesterr}) and unlabeled data, and outlines a prediction
algorithm that achieves this bound.  The intuition behind the
algorithm is that the aggregated prediction on unlabeled examples
should be between $-1$ and $+1$ (see Fig. \ref{fig:predscorefunc}). The "muffling"
behavior occurs when the aggregate prediction on an unlabeled
examples is outside this range. One can represent muffling by
assigning to the unlabeled example a hallucinatory label that is the
opposite of the predicted label.

We use this muffling principle to devise several simple algorithms, 
notably a sequential scheme ($\algname$) that incorporates ensemble classifiers one at a
time into an aggregated classifier.  At every step, it chooses a new
classifier that tends to \emph{disagree} with the current majority
opinion on examples in the unlabeled set.

The rest of the paper is organized as follows. In
Section~\ref{sec:boundtesterr}, we briefly review the theory presented
in \cite{BF15}. In Section~\ref{sec:marvin} we describe the algorithm
$\algname$. In Section~\ref{sec:mower} we describe the algorithm $\hcname$, 
a way of exploiting the partitionings used by ensemble classifiers like decision trees.
Section \ref{sec:experiments} contains a comparative experimental evaluation of our
algorithm on a number of datasets. 
In Section \ref{sec:disc} we draw some conclusions from the experiments,
and in Section~\ref{sec:relwork} we make connections to past and
future work.


\section{Setup: Minimizing Worst-Case Test Error}
\label{sec:boundtesterr}

In this paper, we study a learning scenario where we have two types of data drawn i.i.d. from the same distribution: 
a labeled set $ L = \{ (x^L_1, y_1^L) , \dots, (x^L_m, y_m^L) \}$ and an unlabeled set $U = \{x^U_1, \dots, x^U_n \} $. 
We have at our disposal an ensemble $\cH$ of predictors, 
and are tasked to classify the unlabeled data $U$ as accurately as possible. 
\footnote{We will see that in our case this transductive setting, measuring performance over known $U$, is essentially equivalent to the statistical learning setting, 
where $L,U$ are i.i.d. and the test data are another i.i.d. sample.}

Write $\text{clip} (x) = \min(1, \max( -1, x))$ and $[n] = \{ 1,2,\dots,n \}$, as well as $a_{n_1: n_2}$ to denote the set $\{a_{n_1}, a_{n_1 + 1}, \dots, a_{n_2} \}$. 
All vector inequalities, as well as functions like $\sgn(\vv)$, are component wise. 



Our setting slightly modifies that of \cite{BF15}, 
considering an ensemble of $p$ classifiers. 
Its predictions on the unlabeled data are denoted by $\vF$:
\begin{equation}
\label{eq:defoff}
\vF = 
 \begin{pmatrix}
   h_1(x_1^U) & \cdots & h_1 (x_n^U) \\
   \vdots    & \ddots &  \vdots  \\
   h_p(x_1^U)  & \cdots &  h_p (x_n^U)
 \end{pmatrix}
 \in [-1, 1]^{p \times n}
\end{equation}
We denote the columns of $\vF$ as $\vx_j^U =
(h_1(x_j^U), \cdots, h_p (x_j^U))^\top$, and the rows as $\vh_i = (h_i(x_1^U), \cdots, h_i (x_n^U))^\top$, omitting the superscript $U$. 
The test set has some binary labels $(y_1; \dots; y_n) \in \{-1,1\}^n$, 
which are unknown to the predictor. 
However, the test labels are allowed to be randomized, 
represented by values in $[-1,1]$ instead of just the two values $\{ -1, 1\}$. 
So it is convenient to write the labels on $U$ as $\vz = (z_1; \dots; z_n) \in [-1,1]^n$. 

The idea of \cite{BF15} is to formulate the ensemble aggregation problem as a
two-player zero-sum game between a predictor and an adversary.
In this game, the predictor is the first player, 
who plays $\vg = (g_1; g_2; \dots; g_n)$, 
a randomized label $g_j \in [-1,1]$ for each example $\{x_j\}_{j=1}^{n}$. 
The adversary is then allowed to set the labels $\vz \in [-1,1]^n$. 
A successful predictor player corresponds to a robust learning algorithm, 
able to generalize as well as possible given its lack of full label information. 

The key point is that when any classifier $i$ is known to perform well to a certain degree on the test data, 
its predictions $\vh_i$ on the test data are a reasonable guide to $\vz$, 
and correspondingly give us information by constraining $\vz$ to be "near" them. 
Each classifier in the ensemble thus contributes to an intersecting set of constraints, 
which interact in ways that depend on the ensemble's test predictions.

To discuss these ideas, suppose the predictor has knowledge of a \emph{correlation vector}
$\vb \in (0, 1]^p$ such that
$
\forall i \in [p] , \; \frac{1}{n} \sum_{j=1}^n h_i (x_j) z_j \geq b_i 
$,
i.e. $ \frac{1}{n} \vF \vz \geq \vb$. 
These $p$ inequalities represent upper bounds on individual classifier error rates, 
which can be estimated from the training set w.h.p. when the training and test data are i.i.d., 
in a standard way also used by ERM \cite{BF15}.
So in our game-theoretic formulation, 
the adversary plays under ensemble error constraints defined by $\vb$. 

The predictor attempts to 
\emph{minimize the worst-case expected loss on the test data} 
(w.r.t. the randomized labeling $\vz$), which we write 
$\ell (\vz, \vg) := \frac{1}{n} \sum_{j=1}^{n} \frac{1}{2} (1 - z_j g_j) $. 
The goal is to drive this loss down to $\approx V$, the best upper bound on error that any predictor can guarantee, given the information in $\vF$ and $\vb$: 
\begin{align}
\label{eq:game1eq}
V &:= \min_{\vg \in [-1,1]^n} \; \max_{\substack{ \vz \in [-1,1]^n , \\ \frac{1}{n} \vF \vz \geq \vb }} \; \ell (\vz, \vg)  
\end{align}
The main result of the prior work \cite{BF15} expresses $V$ and the optimal predictor strategy 
$\displaystyle \vg^* $, 
which achieves the optimum of \eqref{eq:game1eq}. 
To state it, define the convex \textbf{potential well} as $\Psi (x) = \max(1, \abs{x})$; 
the \textbf{slack function} as
$\gamma (\sigma) := \gamma_{\vb} (\sigma) := - \ip{\vb} {\sigma} + \frac{1}{n} \sum_{j=1}^n \Psi ( \ip{\vx^U_{j}} {\sigma} )$; 
and the \textbf{optimal weights} $ \sigma^* := \argmin_{\sigma \geq 0^p} \gamma (\sigma)$. 
\footnote{Here $\ip{\cdot}{\cdot}$ denotes the usual inner product in $\RR^p$ -- 
as we move to a setting where the ensemble grows with time $t$, it will denote an inner product over the ensemble classifiers learned so far, 
and will be clear from context.}
Then the semi-supervised aggregation game of \eqref{eq:game1eq} has a conveniently expressible solution. 

\begin{theorem}[\cite{BF15}]
\label{thm:gamesolngen}
The minimax value of the game \eqref{eq:game1eq} is 
\begin{align*}
V = \frac{1}{2} \min_{\sigma \geq 0^p} \lrb{ - \ip{\vb}{\sigma} + \frac{1}{n} \sum_{j=1}^n \Psi ( \ip{\vx_{j}^U}{\sigma} ) } 
= \frac{1}{2} \min_{\sigma \geq 0^p} \gamma (\sigma)
:= \frac{1}{2} \gamma (\sigma^*)
\end{align*}
The minimax optimal predictions for all $j \in [n]$ are 
$g_j^* := [\vg (\sigma^*)]_j := \clip (\ip{\vx_{j}^U}{\sigma^*})$. 
\end{theorem} 

%


This suggests that given any ensemble, we should try to play $\vg^*$ to perform well on $U$, 
finding the $\sigma^*$ that minimizes the slack function $\gamma (\cdot)$ and then playing $\vg (\sigma^*)$. 
We can approximately optimize to find $\sigma \approx \sigma^*$, in which case predicting with $\vg (\sigma)$ is near-optimal (\cite{BF15}). 
This is a semi-supervised alternative to the common supervised learning principle of empirical risk minimization (\cite{V82}), 
in which a bound on training error is minimized. 

The minimax predictor takes an easily interpretable and convenient form. 
For each $x_j^U \in U$, it only depends on the \textbf{score} $\ip{\vx_{j}^U}{\sigma^*}$ of $x_j^U$ 
with respect to the weights $\sigma^*$. 
The \textbf{margin} (defined as $\abs{\text{score}}$) can be interpreted as a notion of confidence, for which this paper provides empirical evidence. 

The average potential $\Psi (\cdot)$ of the unlabeled data regularizes the problem 
by encouraging us to put weight on classifiers which disagree, so that the margin stays low. 
Thm. \ref{thm:gamesolngen} provides a direct proof that this strategy generalizes well by directly addressing test error, 
even though it contrasts starkly with \emph{max-margin} approaches known to generalize in fully supervised settings (\cite{SFBL98}). 

We refer to this minimax framework (of \cite{BF15}) as \emph{muffled learning} 
to emphasize its learning principle of actively distrusting overly confident predictions. 
As a consequence of Thm. \ref{thm:gamesolngen}, 
it always performs at least as well as any single classifier when $\vb$ is estimated accurately -- in other words, unlabeled data do not hurt (\cite{BF15}). 
We follow previous work (\cite{BF15, BF15b}) in emphasizing that while the transductive setting is convenient for a clean muffled formulation, 
in this case it is not a restrictive assumption for high $n$, since the data are i.i.d. (see Appendix \ref{sec:minibatch} for details). 

For the remainder of this paper, we investigate ways to minimize $\gamma (\cdot)$ generally (for any ensemble) and practically. 
All of these are algorithms to find a weight vector $\sigma$ that leads to a good predictor, and to learn the associated ensemble. 
So our algorithms always inherit the aforementioned prediction-based advantages of the muffled learning framework.


\section{An Algorithm for Incrementally Aggregating Classifiers}
\label{sec:marvin}

Directly minimizing the slack function with an ensemble generated a priori like a random forest can enjoy some practical success, 
but it has been reported (\cite{BF15b}) be too conservative, because the bound $V$ on error is too loose. 
However, adding more classifiers to any ensemble can only lower its $V$, because $\vz$ is at least as constrained after the addition. 
Therefore, a natural strategy to mitigate the bound's looseness is to call upon a larger ensemble. 

So we elect to build our predictor incrementally, from classifiers in a possibly infinite ensemble $\cH$. 
Supervised boosting algorithms have long (\cite{F95}) done this efficiently by accessing a learning algorithm that the booster calls as needed, 
to generate classifiers one at a time. 
We also use such a learner as a subroutine -- it returns a classifier from $\cH$ that approximately minimizes error on its input among $h \in \cH$. 
This is efficiently implemented for many hypothesis classes, like decision trees, linear classifiers, and other supervised learning approaches -- 
our method is capable of using any of these. 

Our algorithm, $\algname$, repeatedly requests the classifier in $\cH$ that minimizes error on inputs comprised of $m+n$ weighted examples: 
the $m$ labeled examples in $L$, and the $n$ unlabeled examples in $U$ with purposefully \emph{hallucinated} labels that change every iteration. 
\footnote{
The actual algorithm run is a minibatch version adapted slightly for the stochastic setting (Appendix \ref{sec:minibatch}).
}


\begin{algorithm}[tp]
   \caption{$\algname$}
   \label{alg:realalg}
\begin{algorithmic}
   \STATE 
   {\bfseries Input:} Size-$m$ labeled set $L$, size-$n$ unlabeled set $U$
   \STATE 
   {\bfseries Initialize weights:} $\sigma_0 = \vzero$, so that $\ip{\vx}{\sigma^{t-1}} = 0$ for all $x \in U$
   \FOR{$t = 1$ {\bfseries to} $T$}
   \STATE 
   \textbf{Hallucinate label} for each $x_j^U \in U$: \quad 
   $\displaystyle \hy_{j}^{t} = - \sgn (\ip{\vx^U_j}{\sigma^{t-1}}) \;\cdot \ifn{ \abs{\ip{\vx^U_j}{\sigma^{t-1}}} \geq 1 }$ 
   \vspace{1mm}
   \STATE 
   \textbf{Find a classifier $h^t \in \cH$} that approximately minimizes weighted error over combined data:
   \begin{align}
   \label{eq:ermcall}
   h^t = \argmax_{h \in \cH} \;\lrb{ \frac{1}{m} \sum_{i=1}^{m} y_i^L h (x_i^L) + \frac{1}{n} \sum_{j=1}^{n} \hy_{j}^{t} h (x_j^U) }
   \end{align}
   \STATE 
   \textbf{Add $h^t$ to predictor} with positive weight $\sigma^{t}$ found by line search (e.g. Appendix \ref{sec:gss})
   \STATE
   \emph{Optional, $\algname$-C:} Total correction -- minimize the slack function over the ensemble so far: $h^1, \dots, h^t$ 
   \STATE
   \emph{Optional, $\algname$-D:} If $h^t$ is a decision tree, add all internal nodes of $h^t$ too before performing total correction (see Sec. \ref{sec:mower}).
   \ENDFOR
   \STATE 
   {\bfseries Output:} Predictor $g_T (\vx) =  \mbox{clip}(\ip{\vx}{\sigma^T})$
\end{algorithmic}
\end{algorithm}

$\algname$ is straightforward to specify (Alg. \ref{alg:realalg}), with no parameters to tune. 
It ignores all currently hedged unlabeled examples because they are already minimizing $\Psi$, 
and sends every clipped unlabeled example to the error-minimizing oracle with a hallucinated label of the minority prediction, 
to encourage its margin towards zero. 
Labeled examples are sent to the oracle unchanged, 
and the data are weighted so that $L$ and $U$ have equal weights when no unlabeled data are hedged (see \eqref{eq:ermcall}).

The $\algname$ update can be seen as greedy coordinate descent on the slack function in the high-dimensional space spanned by $\cH$ (Appendix \ref{sec:marvinderiv}). 
This dimensionality is a thorny theoretical issue, so that even though the slack function is convex, 
a practical step size schedule is not easily understood using optimization-based analysis of coordinate descent. 
Even the step size's proportionality constant is of great importance in ensuring that the method converges quickly, 
and a good choice depends on the interactions between dimensions (ensemble classifiers) in complex ways. 
All these considerations motivate us to use line search to find the appropriate step size; 
this is crucial to achieving quick convergence and enabling our total correction results, with details in the appendices.

Another way to improve performance that we experiment with follows the example of \emph{totally corrective} algorithms for supervised boosting (\cite{WLR06}). 
After adding each new ensemble classifier, this approach minimizes the objective function over the entire cumulative ensemble so far. 
It is especially appropriate in our case because the slack function is convex, so efficient optimization methods are guaranteed to make progress. 
We return to this idea in Sec. \ref{sec:experiments}, where we implement $\algname$-C, the totally corrective version of $\algname$.


\section{Maximizing the Performance of an Ensemble of Trees}
\label{sec:mower}

$\algname$ addresses the central issue of learning an ensemble to aggregate, while simultaneously learning the aggregation function also by minimizing the slack function $\gamma(\sigma)$. 
To date, the only other work that has attempted to empirically minimize the slack function is the recent paper \cite{BF15b}, 
whose idea applies to ensembles of decision trees or other partitioning classifiers. 
They augment the ensemble with \emph{specialists} constructed from the leaves of the trees
each of which predicts only on the data falling into it, contributing local information about the true labels. 
The work \cite{BF15b} optimizes using standard gradient descent without line search, 
give evidence that the benefits of such partitioning specialists may complement sequential boosting-type procedures, 
and ultimately pose the fusion of the two approaches as an open problem.

\begin{figure*}[tp]
\vskip -0.1in
\begin{center}
\centerline{\includegraphics[width=\linewidth,height=115pt]{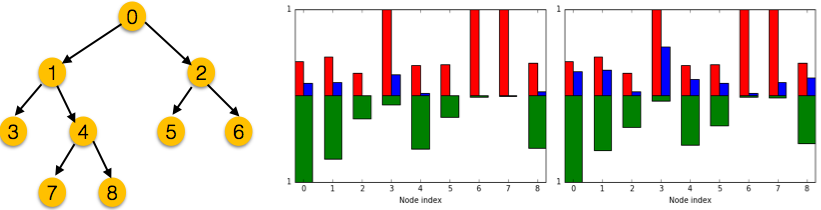}}
\label{icml-historical}
\end{center}
\vskip -0.25in
\caption{Effect of Wilson score interval (Sec. \ref{sec:mower}) for measuring $\vb$, on a decision tree with 9 internal nodes (left). In the middle, using 100 labeled data: for each internal node, green bar is fraction of labeled data falling into that node; red bar is the plugin estimate of $\vb$; blue bar is Wilson lower bound, calculated from values of green and red bars. Right: as middle, but with 400 labeled data.}
\label{fig:wilson_tree}
\end{figure*}

We address this problem, extending the idea of \cite{BF15b} to handle all internal nodes, not just the potentially prohibitively small leaves.
So we need calculate the components of $\vb$, i.e. the errors of all the specialists representing internal nodes, simultaneously. 
These errors are close to their estimates from the labeled data with high probability -- 
this \emph{uniform convergence} of the estimates is deeply studied in learning theory (\cite{V82}). 
The theoretical uniform bounds are too loose for direct use, though, 
so we upper-bound each individually with some very high confidence (e.g. 99.9\%); 
by uniform convergence, this probably constitutes a valid uniform bound on the vector $\vb$.

For each node, we are estimating a binomial proportion (say $p$, using an estimate $\hat{p}$); 
the natural option for this is Wald's confidence interval with width $\sqrt{\frac{\hat{p} (1 - \hat{p})}{m}}$. 
However, this fails to provide adequate coverage in two regimes of interest in decision tree partitionings: 
a small number of labeled data falling into a leaf, 
and very skewed leaves ($\hat{p} \approx 0$ or $1$). 

To maintain coverage of our interval in these situations, we calculate $\vb$ using the lower bound provided by Wilson's score interval (\cite{W27}) for each node of the tree. 
This follows accepted practice for estimating $p$ in the aftermentioned regimes of interest (\cite{BCD01}). 
Figure \ref{fig:wilson_tree} depicts the effect of using Wilson's interval, even on small pure leaves -- it implements nonuniform shrinkage of all errors towards $1/2$ 
to ensure good coverage, 
in keeping with the conservatism of muffling. 
The only parameter here is the confidence level (i.e. the allowed probability of failure) -- 
a higher such probability makes the prediction more aggressive, 
resulting in most of the internal nodes getting "mowed down" to a Wilson lower-bound of $0$, as seen in Figure \ref{fig:wilson_tree}. 
We call the resulting algorithm $\hcname$, minimizing the slack function over the random forest trees combined with all their internal nodes. 
See Sec. \ref{sec:genbeh} for a full specification of the algorithm and of Wilson's score interval.

We find that line search and Wilson's interval are crucially important to our empirical performance. 
Line search results in significant improvements over SGD with a stepsize schedule, even without any additional specialist nodes, 
surprising in light of the reports of this strategy's ineffectiveness in \cite{BF15b}. 
To highlight this, in addition to $\hcname$ we implement a simplified algorithm we call $\hcname$-1, which simply minimizes the slack function using the complete random forest trees, without augmenting with any specialists. 

\begin{figure*}[tp]
\vskip -0.1in
\begin{center}
\centerline{\includegraphics[width=\linewidth,height=140pt]{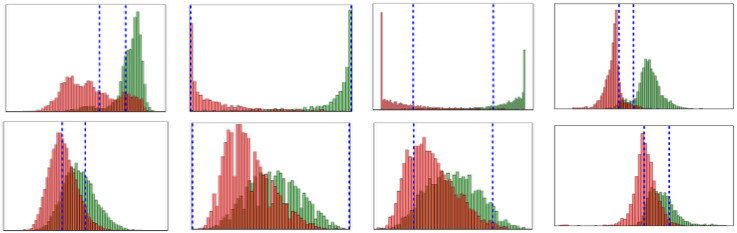}}
\label{icml-historical}
\end{center}
\vskip -0.25in
\caption{Score distributions of four different classifiers in the columns from left to right -- respectively logistic regression, random forests, $\hcname$-1, and $\hcname$. 
Top row: covtype dataset. Bottom row: ssl-secstr. Blue dashed lines are at scores of $\pm 1$, the inflection points of $\Psi$. }
\label{fig:genplots}
\end{figure*}

Following the specialist formulation of \cite{BF15b}, we are adding specialists of many various sizes representing variation at many scales, 
represented by a different scaling factor for each column of $\vF$, so that dimensions are "unnormalized" by design. 
An open problem of \cite{BF15b} is to use second-order or other convex optimization methods to continue to make progress despite such multiscale issues; 
but we believe our approach of first-order line search works satisfactorily here, and it is very efficient (Appendix \ref{sec:gss}).
Fig. \ref{fig:genplots} shows the effect of $\hcname$ and $\hcname$-1 for a couple of datasets; 
the muffling effect of $\Psi$ (recall Fig. \ref{fig:predscorefunc}) is readily apparent, 
particularly when specialist knowledge is incorporated ($\hcname$, right column of Fig. \ref{fig:genplots}).

\section{Empirical Results}
\label{sec:experiments}

We now turn to implementing $\algname$ and $\hcname$ on a variety of datasets. 
We summarize the algorithms we implement; the first four have each been described in previous sections, 
and $\algname$-D combines the two ideas.

\begin{itemize}
\item
$\hcname$-$1$ -- Minimize $\gamma (\sigma)$ using just whole RF trees, and none of their specialists. 
\item
$\hcname$ -- Add all internal nodes of RF trees as specialists, minimize slack function.
\item
$\algname$ -- Add one (non-specialist) tree at a time.
\item
$\algname$-C -- Like $\algname$, but with total correction each timestep.
\item
$\algname$-D -- Similar to $\algname$-C (with total correction), but adding a (non-specialist) tree \emph{and its internal nodes} each timestep, like $\hcname$.
\end{itemize}

Our implementations of our new algorithms are in Python, and use the open-source package \texttt{scikit-learn}. 
We restrict the labeled data available to the algorithm by various orders of magnitude when feasible, to explore its effect. 
Unused labeled examples are combined with the test examples (and the extra unlabeled set, if any is provided) 
to form the set of unlabeled data used by the algorithm. All algorithms are run with 100 base unregularized decision trees as ensemble classifiers where applicable. 
Class-imbalanced and noisy datasets are included, so that AUC is an appropriate measure of performance. 
Results and 95\% confidence intervals are given from 20 Monte Carlo trials (details in appendix), each starting with a different random subsample of labeled data. 
Further information\footnote{Our code is available at \url{https://github.com/aikanor/marvin}.} on the sources of the datasets can be found in the appendices (and in \cite{BF15b}, 
where many of them were used). 
\vspace{0.5em}

\begin{table}[tp]
\centering
\rotatebox{90}{
\scriptsize
\setlength\tabcolsep{1.5pt}
\begin{tabular} 
{| 
M{0.09\linewidth} || M{0.05\linewidth} 
|| M{0.065\linewidth} || M{0.065\linewidth} |  M{0.065\linewidth} || M{0.065\linewidth} || M{0.065\linewidth} || M{0.065\linewidth} |||| M{0.065\linewidth} || M{0.065\linewidth} || M{0.065\linewidth} || M{0.065\linewidth}
|} 
\hline
Dataset & \# labeled & \textsc{Hedge}-\textsc{Mower}-$1$ & \textsc{Hedge}-\textsc{Mower} & \# relevant nodes / \# total nodes per tree & $\algname$ & $\algname$-C & $\algname$-D & RF &Ada-Boost & \hspace{-0em}LogitBoost & LR \\ \hline \hline
 \multirow{3}{*}{\texttt{kagg-prot}} 
 & 100  & 0.58 $\pm$ 0.12 & 0.66 $\pm$ 0.08 & 2 / 12 & 0.65 $\pm$ 0.09 & 0.66 $\pm$ 0.08 & 0.65 $\pm$ 0.12 & 0.64 $\pm$ 0.04 & 0.65 $\pm$ 0.05 & 0.66 $\pm$ 0.06 & 0.64 $\pm$ 0.04  \\ \cline{2-12}
 & 500 & \textbf{0.77 $\pm$ 0.05} & 0.76 $\pm$ 0.05 & 13 / 50 & \textbf{0.80 $\pm$ 0.02} & \textbf{0.83 $\pm$ 0.03} & 0.71 $\pm$ 0.04 & 0.72 $\pm$ 0.03 & 0.72 $\pm$ 0.02 & 0.74 $\pm$ 0.01 & 0.70 $\pm$ 0.02  \\ \cline{2-12}
 & 2500 & 0.83 $\pm$ 0.03 & 0.83 $\pm$ 0.02 & 61 / 242 & 0.79 $\pm$ 0.04 & 0.83 $\pm$ 0.04 & 0.81 $\pm$ 0.05 & 0.80 $\pm$ 0.04 & 0.77 $\pm$ 0.06 & 0.79 $\pm$ 0.06 & 0.75 $\pm$ 0.04  \\ \hline \hline
 \multirow{3}{*}{\texttt{covtype}} 
 & 1K  & 0.81 $\pm$ 0.02 & \textbf{0.82 $\pm$ 0.02} & 23 / 131 & \textbf{0.82 $\pm$ 0.02} & \textbf{0.84 $\pm$ 0.05} & \textbf{0.88 $\pm$ 0.04} & 0.766 $\pm$ 0.014 & 0.729 $\pm$ 0.015 & 0.762 $\pm$ 0.014 & 0.715 $\pm$ 0.015 \\ \cline{2-12}
 & 10K & 0.871 $\pm$ 0.005 & 0.870 $\pm$ 0.015 & 254 / 1103 & 0.942 $\pm$ 0.004 & \textbf{0.969 $\pm$ 0.005} & 0.956 $\pm$ 0.018 & 0.848 $\pm$ 0.006 & 0.738 $\pm$ 0.01 & 0.794 $\pm$ 0.006 & 0.734 $\pm$ 0.009 \\ \cline{2-12}
 & 100K & 0.941 $\pm$ 0.004 & 0.891 $\pm$ 0.005 & 1963 / 8543 & 0.961 $\pm$ 0.005 & \textbf{0.979 $\pm$ 0.003} & \textbf{0.974 $\pm$ 0.003} & 0.926 $\pm$ 0.006 & 0.749 $\pm$ 0.001 & 0.796 $\pm$ 0.002 & 0.769 $\pm$ 0.002 \\ \hline \hline
 \multirow{3}{*}{\texttt{ssl-secstr}} 
 & 100 & 0.58 $\pm$ 0.06 & 0.55 $\pm$ 0.11 & 2 / 18 & 0.61 $\pm$ 0.05 & 0.58 $\pm$ 0.03 & 0.64 $\pm$ 0.06 & 0.55 $\pm$ 0.03 & 0.55 $\pm$ 0.03 & 0.55 $\pm$ 0.04 & 0.56 $\pm$ 0.02 \\ \cline{2-12}
 & 1K  & \textbf{0.70 $\pm$ 0.04} & \textbf{0.68 $\pm$ 0.01} & 14 / 164 & 0.64 $\pm$ 0.02 & 0.64 $\pm$ 0.02 & \textbf{0.66 $\pm$ 0.03} & 0.64 $\pm$ 0.02 & 0.63 $\pm$ 0.02 & 0.64 $\pm$ 0.01 & 0.64 $\pm$ 0.01 \\ \cline{2-12}
 & 10K  & 0.729 $\pm$ 0.006 & \textbf{0.741 $\pm$ 0.006} & 173 / 1630 & 0.680 $\pm$ 0.005 & 0.70 $\pm$ 0.01 & 0.673 $\pm$ 0.010 & 0.673 $\pm$ 0.004 & 0.638 $\pm$ 0.009 & 0.673 $\pm$ 0.006 & 0.701 $\pm$ 0.002 \\ \hline \hline
 \multirow{2}{*}{\texttt{adult}} 
 & 100 & \textbf{0.78 $\pm$ 0.08} & \textbf{0.81 $\pm$ 0.03} & 4 / 16 & 0.73 $\pm$ 0.03 & \textbf{0.77 $\pm$ 0.02} & 0.72 $\pm$ 0.05 & 0.65 $\pm$ 0.09 & 0.68 $\pm$ 0.06 & 0.66 $\pm$ 0.07 & 0.68 $\pm$ 0.08 \\ \cline{2-12}
 & 1K & \textbf{0.84 $\pm$ 0.01} & \textbf{0.85 $\pm$ 0.02} & 29 / 126 & 0.75 $\pm$ 0.02 & 0.78 $\pm$ 0.03 & 0.78 $\pm$ 0.04 & 0.72 $\pm$ 0.03 & 0.73 $\pm$ 0.02 & 0.73 $\pm$ 0.03 & 0.75 $\pm$ 0.03 \\ \hline \hline
  \multirow{2}{*}{\texttt{ssl-text}} 
 & 100 & 0.68 $\pm$ 0.05 & 0.65 $\pm$ 0.09 & 5 / 24 & 0.68 $\pm$ 0.02 & 0.73 $\pm$ 0.08 & 0.69 $\pm$ 0.08 & 0.64 $\pm$ 0.09 & 0.65 $\pm$ 0.05 & 0.64 $\pm$ 0.06 & 0.66 $\pm$ 0.16 \\ \cline{2-12}
 & 1K & 0.85 $\pm$ 0.02 & 0.88 $\pm$ 0.05 & 54 / 237 & 0.81 $\pm$ 0.01 & 0.86 $\pm$ 0.05 & 0.89 $\pm$ 0.06 & 0.83 $\pm$ 0.07 & 0.80 $\pm$ 0.04 & 0.82 $\pm$ 0.07 & 0.86 $\pm$ 0.03 \\ \hline \hline
 \multirow{3}{*}{\texttt{kagg-cred}} 
 & 1K & \textbf{0.74 $\pm$ 0.03} & \textbf{0.75 $\pm$ 0.04} & 21 / 49 & 0.59 $\pm$ 0.02 & \textbf{0.66 $\pm$ 0.06} & \textbf{0.70 $\pm$ 0.10} & 0.56 $\pm$ 0.03 & 0.59 $\pm$ 0.02 & 0.59 $\pm$ 0.03 & 0.53 $\pm$ 0.05 \\ \cline{2-12}
 & 10K & \textbf{0.75 $\pm$ 0.04} & \textbf{0.73 $\pm$ 0.02} & 144 / 356 & 0.62 $\pm$ 0.02 & \textbf{0.75 $\pm$ 0.02} & \textbf{0.72 $\pm$ 0.03} & 0.581 $\pm$ 0.014 & 0.588 $\pm$ 0.015 & 0.590 $\pm$ 0.013 & 0.52 $\pm$ 0.02 \\ \cline{2-12} 
 & 100K & \textbf{0.74 $\pm$ 0.01} & \textbf{0.74 $\pm$ 0.09} & 1404 / 4015 & 0.62 $\pm$ 0.02 & \textbf{0.76 $\pm$ 0.02} & \textbf{0.74 $\pm$ 0.04} & 0.588 $\pm$ 0.005 & 0.591 $\pm$ 0.007 & 0.595 $\pm$ 0.005 & 0.519 $\pm$ 0.008 \\ \hline \hline
 \multirow{2}{*}{\texttt{cod-rna}} 
 & 1K & 0.943 $\pm$ 0.015 & \textbf{0.969 $\pm$ 0.004} & 31 / 80 & 0.929 $\pm$ 0.008 & 0.94 $\pm$ 0.02 & \textbf{0.95 $\pm$ 0.02} & 0.87 $\pm$ 0.03 & 0.90 $\pm$ 0.02 & 0.90 $\pm$ 0.02 & 0.87 $\pm$ 0.02 \\ \cline{2-12} 
 & 10K & \textbf{0.973 $\pm$ 0.006} & \textbf{0.967 $\pm$ 0.008} & 205 / 532 & 0.96 $\pm$ 0.01 & \textbf{0.979 $\pm$ 0.004} & \textbf{0.973 $\pm$ 0.007} & 0.93 $\pm$ 0.02 & 0.91 $\pm$ 0.02 & 0.935 $\pm$ 0.016 & 0.92 $\pm$ 0.02 \\ \hline \hline 
 \multirow{3}{*}{\texttt{SUSY}} 
 & 1K & \textbf{0.82 $\pm$ 0.02} & \textbf{0.830 $\pm$ 0.015} & 25 / 82  & 0.78 $\pm$ 0.02 & 0.76 $\pm$ 0.04 & 0.78 $\pm$ 0.02 & 0.771 $\pm$ 0.006 & 0.769 $\pm$ 0.009 & 0.771 $\pm$ 0.005 & 0.775 $\pm$ 0.006 \\ \cline{2-12}
 & 10K  & \textbf{0.837 $\pm$ 0.005} & \textbf{0.829 $\pm$ 0.009} & 187 / 717 & 0.791 $\pm$ 0.010 & 0.818 $\pm$ 0.005 & 0.801 $\pm$ 0.005 & 0.784 $\pm$ 0.003 & 0.777 $\pm$ 0.006 & 0.788 $\pm$ 0.003 & 0.779 $\pm$ 0.002  \\ \cline{2-12} 
 & 100K  & \textbf{0.849 $\pm$ 0.004} & 0.833 $\pm$ 0.007 & 1547 / 7185 & 0.816 $\pm$ 0.009 & \textbf{0.84 $\pm$ 0.02} & \textbf{0.82 $\pm$ 0.04} & 0.797 $\pm$ 0.003 & 0.797 $\pm$ 0.005 & 0.791 $\pm$ 0.003 & 0.779 $\pm$ 0.002 \\ \hline \hline
\end{tabular}
}
\caption{Area under ROC curve for $\algname$ and $\hcname$ variants, and supervised ensemble algorithms, 
all run with/to 100 ensemble classifiers unless otherwise stated. 
All $\algname$ variants are run with unregularized decision tree weak learners. 
The "\# relevant nodes..." column refers to the average number of internal nodes not mowed down by Wilson's interval 
when running the "$\hcname$" column, 
as a fraction of the total number of internal nodes. 
95\% confidence intervals indicated using 20 Monte Carlo trials, with best algorithm(s) for each row in bold. }
\label{tab:allauc}
\end{table}



We compare our algorithms' performance to that of standard supervised ensemble algorithms under the same conditions -- AdaBoost and LogitBoost, 
random forests (100 trees, default parameters) as a high-performance supervised ensemble algorithm, and logistic regression. 
We find that one or more of our new algorithms is sufficient to achieve significant improvements over the baselines in all cases. 
We further discuss this, and Table 1, in Sec. \ref{sec:disc}.

Many of our datasets are large enough that $U$ will not fit in memory, 
making the batch boosting method impractical. 
However, there is a fairly straightforward minibatch remedy: 
store only a fixed-size minibatch of unlabeled examples, and periodically replace this batch 
(or similar, e.g. in a streaming setting, replace a randomly selected example in the batch with each new example that arrives). 
This is explained in Appendix \ref{sec:minibatch}.


The only tuning done for the new algorithms is of the Wilson failure probability (details in Appendix \ref{sec:genbeh}). 
This applies to all our new algorithms, because all use line search, which requires $\vb$ 
(just one component at a time for $\algname$, and many at once for the other four algorithms). 
The situation is particularly complex for $\algname$-D, which needs enough labeled data to estimate $\vb$ for a growing ensemble including many specialists. 
It is certainly possible that further parameter tuning will lead to better performance in future, but we aim to highlight the approaches' 
simplicity and generality in this initial evaluation.


\section{Discussion}
\label{sec:disc}

The results of Table 1 show that we achieve significant improvement over the baselines in all cases. 
The situation is more unclear when choosing between the new algorithms, an exciting source of future open problems. 
However, we can still deduce some statements and recommendations from Table 1. 

For a combination of simplicity, speed, and good performance with low variance, we recommend $\algname$-C. 
It only adds one new classifier per iteration, 
and its results dominate $\algname$ across experiments; total correction with line search is effective on the convex slack function. 
As the culmination of our ideas in this paper, $\algname$-D might be expected to perform best overall. 
This is possibly true for many datasets, but we cannot conclude this in general, because $\algname$-D often has high variance. 
We believe this has to do with the complex way in which it uses labeled data, both online and to estimate specialist errors. 
Further exploration appears warranted, because such optimization was out of our scope here.


We find that the algorithms here converge quickly in a number of ways. 
Our results typically can be achieved with a small fraction of the unlabeled data available; 
beyond this point, we believe that there is a statistical bottleneck in estimating the first term of the slack function, involving $\vb$. 
In addition, the table makes clear the profound effect of Wilson's interval on $\hcname$, which uses all internal nodes. 
Other heuristics, like selecting just the top $k$ nodes by Wilson score for some $k$, perform almost as well (not shown). 

All this makes the final decision rule of our algorithms essentially a thresholded linear combination of a few white-box tree learners, 
which has the advantageous side effect of being nicely interpretable - the score is just an additive combination of specialist rules, each of which can be written as a decision rule 
(involving both the asleep/awake status and predictions when awake), similar to alternating decision trees \cite{FM99} or similar tree ensembles.

\section{Related and Future Work}
\label{sec:relwork}

Semi-supervised learning has been an active area of research over the last decade (\cite{CSZ06}), 
mostly involving graph-based methods like label propagation that operate pairwise on the data, and also including the transductive SVM \cite{J99} and other algorithms \cite{ZG09}. 
These generally try to locate the decision boundary at low-density regions of the unlabeled data (\cite{CZ05});  
when formulated as max-margin methods, they stand in stark contrast to the muffled min-margin idea for generalization. 
The labels typically agree with the labeled data and some type of unlabeled cluster structure used as a regularizer (\cite{BNS06}), 
while our regularizer is in the same spirit but encourages the opposite muffling behavior. 
Other more coarse-grained methods using discriminative statistics \cite{CSZ06, QSCL09} are more in the spirit of our algorithms.

Semi-supervised algorithms for boosting have previously drawn some attention for their applications, 
notably in \cite{GLB08, KJJL09}, which also hallucinate labels over the unlabeled data; 
but they do not use the muffling framework. 
The only practical work which does is the aforementioned method of \cite{BF15b}, 
and we directly address a main open problem posed in that paper, about combining specialist information with the incremental aggregation idea of boosting. 
Another fascinating open problem, building further on these ideas, 
is how to target areas of the space with specialist classifiers as part of the incremental process, 
rather than just using the specialists provided by decision trees.

This paper is related to the significant existing supervised boosting literature \cite{SF12}. 
Such algorithms concern the incremental classifier aggregation idea, 
and generally attempt to minimize some convex upper bound on error on $L$. 
It would be of interest to incorporate other notions from boosting theory, like weak learnability, into our framework, 
or investigate if they are even necessary.

Our understanding of the generalization behavior of the muffling framework is still just beginning, though it is clear that estimation of $\vb$ is heavily involved. 
There is already a theoretical connection established between this estimation, generalization, and classifier complexity as measured by $\vnorm{\sigma}_1$, 
which relates to $L_{\infty}$ norm constraints on the adversary in the minimax formulation (\cite{BF15c}), 
and building on this could yield fruitful practical insights. 

Finally, we plan to explore practical applications at larger scale to investigate the space of ensembles that can be aggregated -- 
for instance, decision trees can be inappropriate in high dimension, and efficient linear classifiers could be used instead. 
Deep learning of features is another possibility we would like to explore with the muffling framework, 
especially in light of the profound and rapidly expanding set of connections between deep and semi-supervised learning (\cite{B09}).

\newpage
\small
\bibliography{nips16SSboost}
\bibliographystyle{unsrt}

\newpage
\appendix

\section{Derivation of the $\algname$ Update Rule}
\label{sec:marvinderiv}

The idea of hallucinating labels on unlabeled data is not new to semi-supervised learning (see Sec. \ref{sec:relwork}). 
But in our case, we can show that $\algname$ is approximately a greedy coordinate descent update on the slack function, with $\vb$ estimated using $L$: 
\begin{align}
\label{eq:avglosstoslack}
\gamma &(\sigma_T)
= - \ip{\vb}{\sigma_T} + \frac{1}{n} \sum_{j=1}^n \Psi \lrp{ \ip{\vx^U_{j}} {\sigma_T} }  \\
&\approx \frac{1}{m} \sum_{j=1}^{m} \lrb{- y_j^L \alpha \sum_{t=1}^T h_t (x_j^L) } + \frac{1}{n} \sum_{j=m+1}^{m+n} \Psi \lrp{ \alpha \sum_{t=1}^T h_t (x_j^U) } \nonumber \\
&= - \sum_{t=1}^T \frac{1}{m} \sum_{j=1}^{m} \lrp{ y_j^L \alpha h_t (x_j^L) } + \frac{1}{n} \sum_{j=m+1}^{m+n} \Psi \lrp{ \alpha \sum_{t=1}^T h_t (x_j^U) } \nonumber
\end{align}
Taking the partial derivative with respect to any $\{h_i\}_{i=1}^p$ and minimizing over $i$ for the steepest descent direction, we get the $\algname$ update.


\section{Generalization and Estimating $\vb$}
\label{sec:genbeh}

Here we expand on the discussions of Section \ref{sec:mower}. 

\subsection{Wilson's Interval}
\label{sec:wilsonspec}

Wilson's score interval is specified as follows, for a binomial proportion. 
Our problem of bounding the error rate of a specialist from data is like determining the unknown bias $p \in [0,1/2]$ of a biased coin that comes up heads ($1$) with probability $p$, 
using $n$ random flips $A_1, \dots, A_n \in \{0, 1\}$ to estimate a high-probability upper bound for $p$. 
In our case, $p$ is the error rate of the specialist, and $n$ the number of labeled data predicted upon by the specialist and used to estimate its error. 

We would like a high-probability upper bound $p_u$ for $p$, with specified failure probability $\alpha \in [0,1]$; 
we wish that $p \leq p_u$ w.p. $1-\alpha$ over the $n$ coin flips (the labeled data). 

The most apparent unbiased estimator of $p$ is $\hat{p} = \frac{1}{n} \sum_{i=1}^n A_i$. 
The commonly used Wald confidence interval uses the fact that $n \hat{p}$ is binomially distributed: 
$n \hat{p} \sim Bin(n, p)$, and plugs in $\hat{p}$ instead of $p$ for the bias of the binomial. 
This results in an upper bound on error of: 
\begin{align*}
p_u = \hat{p} + \hat{\sigma} z_{\alpha} \qquad\mbox{where}\qquad \hat{\sigma} = \sqrt{\frac{\hat{p} (1 - \hat{p} )}{n}}
\end{align*}
where $z_{\alpha}$ is the $(1-\alpha)$ quantile of the standard normal distribution. 

Wilson's interval instead shows shrinkage toward $\frac{1}{2}$, along with additive bias and variance corrections:
\begin{align*}
p_u = \tilde{p} + \tilde{\sigma} z_{\alpha} \qquad\mbox{where}\qquad \tilde{p} = \frac{\hat{p} + \frac{z_{\alpha}^2}{2n} }{1 + \frac{z_{\alpha}^2}{n}}  \quad, \qquad 
\tilde{\sigma} = \frac{ \sqrt{\frac{\hat{p} (1 - \hat{p} )}{n} + \frac{z_{\alpha}^2}{4 n^2} } }{1 + \frac{z_{\alpha}^2}{n}}
\end{align*}
The interval is derived by considering the behavior of the exact binomial tail for low $n$, the regime in which all approximations to such an exact tail fail. 
Our usage often requires this, particularly when considering leaves of an unregularized decision tree as specialists.
Wilson's interval is therefore numerically appropriate for our usage; further discussions on numerical stability can be found in the excellent overview of \cite{BCD01}.

\subsection{Other Details}

We selected the allowed failure probability for Wilson's score interval by cross-validating among only a few values (or not at all), 
depending on the number of labeled data $m$: $\lrsetb{0.01}$ for $m=1K$, $\lrsetb{0.001, 0.005}$ for $m=10K$, $\lrsetb{0.001}$ for $m=100K$. 
These work across datasets, and across our new algorithms, to give significant performance improvements. 
Small $m$ values are more problematic to deal with -- the predictor has naturally higher variance -- so for those experiments we choose from $\lrsetb{0.003, 0.01, 0.03, 0.1}$.

All our algorithms put labeled data to two different uses: training the ensemble itself (RF or incrementally), 
and estimating $\vb$. 
For $\hcname$ and $\hcname$-1, we recommend using less training data in general when $m$ is large (as supervised generalization anyway limits the predictive power of each ensemble classifier), 
in favor of more accurately estimating $\vb$. 
We use 1/4 of the labeled data to train and 3/4 to measure $\vb$; this works well across all experiments. 

For $\algname$,

%


\begin{algorithm}[tp]
   \caption{$\hcname$}
   \label{alg:realalg}
\begin{algorithmic}
   \STATE 
   {\bfseries \textsc{Input}:} Size-$m$ labeled set $L$, size-$n$ unlabeled set $U$, number of trees $p$, 
   Wilson interval tail probability $\alpha$
   \STATE 
   {\bfseries Partition} $L$ at random into sets $L_1$ (of size $\frac{m}{4}$) and $L_2$
   \STATE 
   \textbf{Train} random forest with $p$ trees using $L_1$
   \STATE
   \textbf{Calculate $\vb$} for the $p$ trees and all their internal nodes using the lower bound of Wilson's interval with tail prob. $\alpha$
   \STATE
   \textbf{Prune $\vb$} to leave only nodes with Wilson lower bounds $\geq 0$ (\emph{Optional: }Further prune, by Wilson scores or otherwise. $\hcname$-1 prunes away all except the $p$ original non-specialist trees. )
   \STATE 
   \textbf{Approximately minimize slack function} using $L_2$ and $U$ to find:
   \begin{align}
   \sigma^o \approx \argmin_{\sigma \geq 0^p} \;\lrb{ - \ip{\vb}{\sigma} + \frac{1}{n} \sum_{j=1}^n \Psi ( \ip{\vx_{j}^U}{\sigma} ) }
   \end{align}
   (Done in this paper with gradient descent using line search. )
   \STATE 
   {\bfseries \textsc{Output}:} Predictor $g (\vx) =  \mbox{clip}(\ip{\vx}{\sigma^o})$
\end{algorithmic}
\end{algorithm}

\section{Implementation Details}

Discuss the minibatch version that uses a constant amount of memory even when unlabeled data won't fit. 
Can do the same when labeled data won't fit.
Unlabeled stride of 100 suffices always, and each iteration is computationally cheap.

Discuss how we choose the unlabeled stride size; otherwise, computation takes longer and convergence may be slower. 
There is a computation-statistics tradeoff here. Total correction is implemented by running stochastic gradient descent for 100 iterations 
after adding each new classifier.

\subsection{Algorithms for Statistical Learning and Streaming Settings}
\label{sec:minibatch}

Many of our datasets are large enough that $U$ will not fit in memory, 
making the batch boosting method impractical. 
However, there is a fairly straightforward minibatch remedy: 
store only a fixed-size minibatch of unlabeled examples, and periodically resample this entire minibatch, or a similar method 
(in a streaming setting, replace a randomly selected example in the batch with each new example that arrives). 

This works because in the muffled formulation, the unlabeled data only enter into the optimization through the average value of the potential well. 
Since the data are i.i.d., this can be estimated with just a small sample, 
so roughly speaking, the transductive setting converges to the i.i.d. setting when $n$ is high (lots of unlabeled data). 
This could conceivably be done for the labeled data as well, but to compare to batch supervised baselines we limit $m$ to fit in our memory in this paper. 

There are potentially situations in which the hierarchical partitioning of the data is known by other means than a decision tree, 
such as an unsupervised clustering method. 
In such cases, the labeled data can exclusively be used to estimate $\vb$ for each specialist, but the specialist predictions are not defined a priori. 
Here it is natural to associate each node with just one prediction: the majority label of the data falling within it. 

We experimented with this method using the partitioning defined by the decision tree (not shown), 
and it performs comparably to or slightly worse than the methods in this paper; 
we believe this is because decision trees are learning a supervised hierarchical partition, 
so the granularity of their predictions is useful. 
Further discussion is outside our scope here, 
but this variant might be better for distributed applications, where each node can store its own one-bit prediction irrespective of the others.

\subsection{Golden Section Line Search for the Step Size}
\label{sec:gss}


We use a modified (memoized) version of golden section search for our line search, 
which is crucial to our algorithms. 
It is described well in the original paper \cite{K53} and commonly in textbooks; 
we use the version given in the \texttt{scipy.optimize} package.

\begin{table}[tp]
\begin{tabular} 
{| 
M{0.15\linewidth} || M{0.12\linewidth} 
|| M{0.15\linewidth} || M{0.10\linewidth} || M{0.30\linewidth} 
|} \hline
Dataset & \# labeled & \# unlabeled & Dim. & Comments \\ \hline \hline
 \multirow{1}{*}{\texttt{kagg-prot}} & 3750 &   & 1776 & Kaggle challenge \cite{kagg2} \\ \hline 
 \multirow{1}{*}{\texttt{ssl-text}} & 1500 &   & 11960 & \cite{CSZ06} \\ \hline
 \multirow{1}{*}{\texttt{kagg-cred}} & 150K &   & 10 & Kaggle challenge \cite{kagg1}; imbalanced (< 10\% positives) \\ \hline
 \multirow{1}{*}{\texttt{adult}} & 32561 &  & 123 & LibSVM \\ \hline
 \multirow{1}{*}{\texttt{covtype}} & 581012 &   & 54 & LibSVM \\ \hline
 \multirow{1}{*}{\texttt{ssl-secstr}} & 83679 & 1189472 & 315 & \cite{CSZ06} \\ \hline
 \multirow{1}{*}{\texttt{cod-rna}} & 59535 train, 271617 test & 157413 & 8 & LibSVM  \\ \hline 
 \multirow{1}{*}{\texttt{SUSY}} & 5M &   & 18 & UCI  \\ \hline
\end{tabular}
\caption{Information about the datasets used. }
\label{tab:allauc}
\end{table}
\vspace{-2mm}

\end{document}